\documentclass{article}

\usepackage[preprint]{neurips_2023_otml}

\usepackage{graphicx}
\usepackage{algorithm}
\usepackage{fnpct}
\usepackage{footnote}
\usepackage{graphics} 
\usepackage{graphicx} 
\usepackage{epsfig} 
\usepackage{times} 
\usepackage{amsmath} 
\usepackage{amssymb}  
\usepackage{tikz}
\usepackage{dirtytalk}
\usetikzlibrary{arrows.meta, positioning, shapes}
\usepackage{url,verbatim,color,comment}
\usepackage[noend]{algpseudocode}
\usepackage{eucal}
\usepackage{amsfonts}
\usepackage{xcolor}
\usepackage{blindtext}
\usepackage{mathtools}
\usepackage{placeins}
\usepackage{fancyhdr}
\usepackage{dirtytalk}

\title{Understanding Reward Ambiguity Through Optimal Transport Theory in Inverse Reinforcement Learning}

\author{%
  Ali Baheri\\
  Department of Mechanical Engineering\\
  Rochester Institute of Technology\\
  Rochester, NY 14623 \\
  \texttt{akbeme@rit.edu} \\
}

\begin{document}

\maketitle

\begin{abstract}

In inverse reinforcement learning (IRL), the central objective is to infer underlying reward functions from observed expert behaviors in a way that not only explains the given data but also generalizes to unseen scenarios. This ensures robustness against reward ambiguity—where multiple reward functions can equally explain the same expert behaviors. While significant efforts have been made in addressing this issue, current methods often face challenges with high-dimensional problems and lack a geometric foundation. This paper harnesses the optimal transport (OT) theory to provide a fresh perspective on these challenges. By utilizing the Wasserstein distance from OT, we establish a geometric framework that allows for quantifying reward ambiguity and identifying a central representation or \say{centroid} of reward functions. These insights pave the way for robust IRL methodologies anchored in geometric interpretations, offering a structured approach to tackle reward ambiguity in high-dimensional settings.

\end{abstract}
\vspace{-3 mm}
\section{Introduction}

In the reinforcement learning (RL), the derivation of reward functions has always been a subject of intricate debates and investigations \cite{amodei2016concrete}. Inverse RL (IRL) emerges as a pivotal methodology in this dialogue, where the aim is to decipher the underlying reward structure from observed expert behavior \cite{ng2000algorithms}. However, one of the intrinsic challenges that plagues IRL is the issue of reward ambiguity. A fundamental realization in IRL is that multiple distinct reward functions can lead to the same or similar expert behaviors. This many-to-one relationship between reward functions and policies complicates the task of pinpointing a unique reward landscape from demonstrations \cite{ng2000algorithms,abbeel2004apprenticeship}.

Attempts to navigate this conundrum have been diverse. The maximum entropy IRL framework, for instance, integrates a probabilistic lens into IRL \cite{ziebart2008maximum}. By postulating that expert demonstrations align not solely with optimality but also with an exponential reward distribution, this approach employs entropy maximization to favor reward functions yielding more uniformly distributed trajectories, serving as an inherent countermeasure to reward ambiguity. Further advancements came from Bayesian IRL, where the uncertainty in reward functions is addressed directly by employing a Bayesian framework \cite{ramachandran2007bayesian,choi2012nonparametric,choi2011map}. Instead of seeking a single reward function, Bayesian IRL estimates a posterior distribution over reward functions given the demonstrations. InfoGAIL, an extension of the GAIL framework, is another noteworthy mention \cite{li2017infogail,ho2016generative}. It incorporates mutual information into the objective, aiming to capture multiple modes of the expert's behavior. By doing so, InfoGAIL effectively disentangles different underlying reasons (latent variables) for the expert's actions, which can be seen as a means to shed light on the ambiguous regions of the reward landscape.

Optimal transport (OT) theory presents a fresh, promising avenue for reward ambiguity problem. The fundamental tenet of OT lies in its ability to measure the \emph{distance} between different probability distributions, or in our context, between various reward functions. This is achieved using a concept called the Wasserstein distance. By understanding how far apart or close together different reward functions are, one can gain clearer insights into the nebulous territory of reward ambiguity that IRL grapples with. The true potential of OT for IRL is its \emph{geometrically-grounded} framework. Where traditional approaches might only offer probabilistic insights, the marriage of IRL and OT provides a more structured, spatial understanding of reward landscapes. Imagine mapping reward functions within a geometric space, where the proximity between points indicates similarity. OT provides the tools to construct, measure, and interpret this space, and in doing so, can potentially revolutionize our comprehension of reward ambiguity. In this paper, our primary contribution is the fusion of the analytical power of OT with the challenges of reward ambiguity inherent in IRL. We advocate for a synthesis, where the discerning capabilities of IRL meet the geometric precision of OT, aiming to deliver a more holistic solution to one of RL's most enduring challenges.
\vspace{-3 mm}
\section{Preliminaries}

OT finds its roots in the study of moving and redistributing masses in the most efficient manner \cite{villani2009optimal,santambrogio2015optimal}. Historically conceived by Gaspard Monge in the late 18th century, its motivations were primarily combinatorial. However, with the reformation of the theory by Kantorovich in the 20th century, it adopted a more functional analysis perspective, making it more broadly applicable \cite{kantorovich2006translocation,kantorovich2006problem}. In the context of OT, we deal with two probability measures, $\mu$ and $\nu$, defined on a common metric space. The foundational goal of OT is to \say{transport} $\mu$ to $\nu$ while minimizing a certain cost associated with moving each infinitesimal unit of mass. This cost is typically governed by a cost function $c$, which dictates the expense of moving a unit mass across a given distance.
Mathematically, the $p$-Wasserstein distance between two probability measures $\mu$ and $\nu$, defined with respect to a ground metric $d$ on the space, is given by:
\begin{equation}
W_p(\mu, \nu)=\left(\inf _{\gamma \in \Gamma(\mu, \nu)} \int d(x, y)^p d \gamma(x, y)\right)^{\frac{1}{p}}
\end{equation}
In this representation, $x$ and $y$ are points within the metric space over which the probability measures $\mu$ and $\nu$ are defined. The function $d(x, y)$ computes the distance between these two points, rooted in the foundational ground metric $d$. Meanwhile, $\Gamma(\mu, \nu)$ signifies the set of all conceivable joint distributions (or transport plans) having $\mu$ and $\nu$ as marginals. The Wasserstein distance's merit is its efficacy to encapsulate the geometric discrepancies between probability measures. This capability has rendered it invaluable across various domains, from image processing to machine learning, economics, and computational biology \cite{ferradans2014regularized,cuturi2013sinkhorn,baheri2023risk,luo2023optimal,galichon2022cupid,schiebinger2019optimal}. In the context of our study, the appeal of OT is derived from its capacity to numerically elucidate disparities between reward function distributions. This offers a unique perspective to examine the intricacies of reward ambiguity inherent in IRL.
\vspace{-3 mm}
\section{Problem Formulation}

IRL revolves around the fundamental task of inferring the most probable reward function that an expert is optimizing, given their behavior. A key challenge in IRL is the ambiguity of reward functions: multiple reward functions might give rise to the same, or nearly identical, expert policies. This set of reward functions is denoted as $R\left(\pi^*\right)$, where $\pi^*$ is the expert's policy. A pertinent factor influencing the dynamics of this ambiguity is the dimensionality of the problem. In our context, the state space is represented as $\mathcal{S}$ and the action space as $\mathcal{A}$. The dimensionality, $d$, encapsulates the total number of state-action pairs and is given by:
\begin{equation}
d=|\mathcal{S}| \times|\mathcal{A}|    
\end{equation}
Higher the dimensionality, the more intricate and vast our reward space becomes, potentially complicating our understanding of $R\left(\pi^*\right)$.
To navigate this complex space and to better understand the distribution and relationships between these reward functions, we turn to OT theory. This theory allows us to map each reward function in $R\left(\pi^*\right)$ to a representation in a space where distances (dissimilarities) between these functions can be meaningfully measured. We term this mapping as $\Phi$, which translates a reward function, $R$, to its representative point, $W$, in the Wasserstein space.
With this transformation, \textbf{our aims become twofold:}

\noindent{\textbf{1. Quantification.} Measure the Wasserstein distance between different representations in the Wasserstein space, $\mathcal{W}\left(W_1, W_2\right)$. This measurement provides insight into the degree of ambiguity among reward functions.

\noindent{\textbf{2. Centroid Identification.} Beyond just understanding the space, it is crucial to pinpoint a central, representative point in this space, a \say{centroid} $W^*$, that stands, on average, closest to all other reward function representations. Mathematically, this can be articulated as:

%
\begin{equation}
W^*=\operatorname{argmin}_{W \in \Phi\left(R\left(\pi^*\right)\right)} \sum_{W_i \in \Phi\left(R\left(\pi^*\right)\right)} \mathcal{W}\left(W, W_i\right)
\end{equation}
This centroid aids in identifying a \say{central} reward function representation, which can serve as a reference or benchmark in the Wasserstein space.

\vspace{-3 mm}
\section{Theoretical Results}

Upon formalizing the inherent challenges of reward ambiguity, particularly in the context of escalating dimensionality, the immediate imperative is to elucidate these complexities with rigorous theoretical underpinnings. The potential of OT, as postulated in the preceding sections, necessitates concrete validation. In this section, a suite of theorems is presented. These results not only affirm our foundational hypotheses but also offer a systematic framework to comprehend and navigate the intricacies of reward function ambiguity in IRL. \footnote{Proofs for the presented theorems can be found in the supplementary material.}

\noindent{\textbf{Theorem 1: Convergence of Inferred Rewards in Wasserstein Space.
}
Let $R_{\text {true }}$ be the true underlying reward function of an expert, and $R_n$ be the inferred reward function based on $n$ expert trajectories. Then, under certain regularity conditions on the environment and expert behavior, as $n \rightarrow \infty$ :
\begin{equation}
W_p\left(R_n, R_{\text {true }}\right) \rightarrow 0
\end{equation}
where $W_p$ denotes the $p$-Wasserstein distance.}}

\noindent{\textbf{Implications.} The theorem explaining the convergence of inferred rewards in the Wasserstein space provides key guarantees for IRL. At its core, it confirms that, given a broad range of expert demonstrations, our estimated reward structures gradually approach the true underlying reward function driving the expert's decisions. This points to a consistent pattern in expert behaviors, addressing the well-known issue in IRL where multiple reward functions could seemingly explain the same expert actions. Furthermore, the theorem's foundation in OT enhances the reliability of IRL, suggesting that even with potential inaccuracies and variability in real-world demonstrations, the geometric and topological insight from the Wasserstein distance offers a way to identify and get closer to the true reward function. Therefore, this convergence theorem not only strengthens the theoretical basis of IRL but also highlights the crucial role of OT in improving the precision of reward predictions based on expert data.

\noindent{\textbf{Theorem 2: Existence and Uniqueness of Optimal Reward.}
Let $\mathcal{R}$ be a compact metric space representing the space of all reward functions, $\pi^*$ be an expert's policy, $\mathcal{S}=\left\{R_1, R_2, \ldots, R_N\right\}$ be a set of inferred reward functions from $\mathcal{R}$, where each $R_i$ induces a policy equivalent to $\pi^*$, and $W_p(\cdot, \cdot)$ be the $p$-Wasserstein distance induced by a ground metric on $\mathcal{R}$. Then: 
\newline
\noindent{\textbf{1. Existence:} There exists a reward function $R_{\text {centroid }} \in \mathcal{R}$ such that
\begin{equation}
R_{\text {centroid }}=\underset{R \in \mathcal{R}}{\operatorname{argmin}} \sum_{i=1}^N W_p\left(R, R_i\right)
\end{equation}
This $R_{\text {centroid }}$ minimizes the total $p$-Wasserstein distance to all reward functions in $\mathcal{S}$.

\noindent{\textbf{2. Uniqueness:}} If the ground metric on $\mathcal{R}$ induces strict convexity, then $R_{\text{centroid}}$ is unique.

\noindent{\textbf{Implications.} The theorem elucidating the existence and uniqueness of the optimal reward representation within the Wasserstein space offers profound insights for the IRL. It posits that amidst the vast space of plausible reward functions that can induce a given expert policy, there emerges a central or \say{centroid} reward function that serves as the most representative embodiment with respect to the Wasserstein distance. This centrality implies a kind of \say{mean} or \say{average} behavior which, in the face of reward ambiguities, can provide a benchmark for the modeling of underlying motivations or preferences. Furthermore, the condition for uniqueness, contingent on strict convexity of the ground metric, reassures that this centroid reward is not only representative but singularly distinctive, thereby eliminating concerns of multiple competing central representations. 

\noindent{\textbf{Theorem 3: Robustness of OT-Integrated IRL to Noisy Demonstrations.}} Let $\epsilon$ be a measure of the noise introduced in the demonstrations such that $\epsilon \geq 0$, with $\epsilon=0$ representing noise-free demonstrations. Then, for every $\epsilon>0$, there exists a constant $C(\epsilon)$ such that:
\begin{equation}
    W_p\left(R^*\left(\pi^*\right), R^*\left(\pi^{* \prime}\right)\right) \leq C(\epsilon)
\end{equation}
This implies that the distance between reward functions inferred from noisy and noise-free demonstrations remains bounded in the Wasserstein space, ensuring robustness of the OT-integrated IRL process to noisy expert demonstrations.

\textbf{Implications.}
The theorem underscoring the robustness of the OT integration in IRL against noisy demonstrations bears significant ramifications for practical implementations. It suggests that even when expert demonstrations are marred by imperfections or inaccuracies – a common occurrence in real-world scenarios – the inferred motivations or reward structures remain steadfastly close to those derived from noise-free demonstrations. This bounded divergence in the Wasserstein space acts as a guarantee of stability, ensuring that the model's interpretation of an expert's intent is not drastically swayed by minor fluctuations or disturbances in the input data. 

\noindent{\textbf{Theorem 4: Impact of Dimensionality on Reward Ambiguity in Wasserstein Space.}}
Given the set $S_d$ of reward functions for dimension $d$, let $\Delta_d$ be the average pairwise $p$-Wasserstein distance between reward functions in $S_d$ defined as
\begin{equation}
\Delta_d=\frac{2}{\left|S_d\right|\left(\left|S_d\right|-1\right)} \sum_{R_i, R_j \in S_d, i<j} W_p\left(R_i, R_j\right)
\end{equation}
Then, as $d$ increases, $\Delta_d$ also increases.
This indicates that as the dimensionality of the state-action space grows, the average pairwise Wasserstein distance between potential reward functions grows as well, suggesting a more dispersed reward landscape and thus intensified reward ambiguity.

\noindent{\textbf{Implications.}
The theorem indicates that higher dimensionality in the state-action space corresponds to a greater average pairwise Wasserstein distance between possible reward functions. This, in turn, points to a more scattered distribution of reward functions in the geometric space. The spread of these reward functions implies that as we consider problems with more dimensions, the ambiguity in determining the \say{true} reward function (among many that can explain the expert behavior) becomes more pronounced.

\vspace{-3 mm}
\section{Discussion}

OT is effective at analyzing and comparing probability distributions by viewing them as comprehensive entities. However, this approach can pose challenges when dealing with reward functions that exhibit multiple modes. A unimodal representation in the Wasserstein space may not capture all the distinct peaks of a multimodal function. As a result, the overall representation might be oversimplified or inaccurate. To capture the diverse peaks of multimodal reward functions, it could be beneficial to represent them using multiple points or clusters in the Wasserstein space. Each cluster could represent a different peak or mode, allowing for a more detailed depiction of the reward function. Yet, this method introduces its own set of challenges, particularly in computation. Using multiple clusters necessitates specialized clustering algorithms suitable for the Wasserstein space, which could increase computational demands. Nonetheless, recent improvements in OT algorithms provide promising solutions to these challenges \cite{peyre2017computational}. While employing multiple clusters enhances the representation's accuracy, it can make the decision-making process more complex. How should an agent prioritize between multiple modes of high reward? Incorporating these insights into actionable policies might require additional layers of analysis or even a hierarchical approach to decision-making.
\vspace{-3 mm}
\section{Conclusion}
\label{sec:conclusion}

IRL faces with issues such as reward ambiguity, where multiple reward functions can justify identical expert behaviors. In this paper, we presented an approach by incorporating optimal transport theory, specifically utilizing the Wasserstein distance. This method brings forth a geometric insight into reward ambiguity, paving the way for a more systematic understanding. While the theoretical underpinnings presented in this work shed light on several facets of the reward ambiguity challenge, a crucial avenue for future research is empirical validation. Rigorous testing and application of our theoretical insights to real-world IRL scenarios will be essential to fully gauge the efficacy and broad applicability of our proposed framework.

\newpage

\bibliographystyle{unsrt}
\bibliography{neurips_2023_otml}

\newpage
\section{Supplementary Material}
\noindent{\textbf{Proof of Theorem 1.}}

\noindent{\textbf{Lemma 1.}} For every trajectory and any inferred reward function $R$, there exists a constant $C$ such that: $|R(s, a)| \leq C$
for all state-action pairs $(s, a)$.

\noindent{\textbf{Proof.} The reward values should be bounded for practical tasks and environments; otherwise, they could lead to unbounded utility values making the problem ill-posed.

\noindent{\textbf{Lemma 2.}} \textit{Consistency of IRL Inference.}
Given the true reward function $R_{\text {true, }}$, as the number of expert trajectories $n \rightarrow \infty$, the expected reward under the inferred policy $\pi_n$ based on $R_n$ converges to the expected reward under the expert's policy $\pi^*$ based on $R_{\text {true }}$.

\noindent{\textbf{Proof.} Let's consider the set of all possible trajectories $\tau$ generated using the expert's policy $\pi^*$. Since the policy is fixed and deterministic, the distribution of trajectories is also fixed, denoted by $p\left(\tau \mid \pi^*\right)$. Given a sufficiently large number $n$, the likelihood of observing a particular trajectory $\tau$ using the inferred policy $\pi_n$ should approach the likelihood of observing $\tau$ using the expert's policy $\pi^*$. Therefore, as $n \rightarrow \infty$ :
\begin{equation}
\mathbb{E}_{\pi_n}\left[R_n(s, a)\right] \rightarrow \mathbb{E}_{\pi^*}\left[R_{\text {true }}(s, a)\right]
\end{equation}

\noindent{\textbf{Proposition.} \textit{Compactness in Wasserstein Space.}
Given a set of inferred reward functions $\left\{R_n\right\}$ based on increasing number of expert trajectories, the sequence is relatively compact in the Wasserstein space.

Proof of Theorem:

From Lemma 1, we know the rewards are bounded. This implies that the set of potential reward functions resides in a compact space. From Lemma 2, as $n \rightarrow \infty$, the difference in expected rewards under the inferred policy and the expert's policy diminishes. Given the compactness of the space of reward functions (from the proposition) and the convergence of expected rewards, it implies that the sequence $\left\{R_n\right\}$ has a convergent subsequence in the Wasserstein space. By the properties of Wasserstein convergence, this convergent subsequence must converge to the true reward function $R_{\text {true }}$ in the Wasserstein metric. Thus, for any $\epsilon>0$, there exists an $N$ such that for all $n>N$ :
\begin{equation}
W_p\left(R_n, R_{\text {true }}\right)<\epsilon
\end{equation}
This implies that as $n \rightarrow \infty, W_p\left(R_n, R_{\text {true }}\right) \rightarrow 0$.

\noindent{\textbf{Proof of Theorem 2.}

\noindent{\textbf{Lemma 1.} The sum of $p$-Wasserstein distances, as a function on $\mathcal{R}$, is lower bounded and continuous.

\noindent{\textbf{Proof.} \textit{Lower Bounded}: By the non-negativity of the Wasserstein distance, the sum is clearly nonnegative and thus lower bounded by 0.
\textit{Continuity}: The Wasserstein distance itself is continuous by its definition, and the sum of continuous functions remains continuous.
Given that $\mathcal{R}$ is compact and the sum of $p$-Wasserstein distances is continuous on $\mathcal{R}$, by the Extreme Value Theorem, the function attains its minimum. Hence, a minimizer $R_{\text {centroid }}$ exists.

\noindent{\textbf{Lemma 2.} If the ground metric space is strictly convex, then the $p$-Wasserstein distance is also strictly convex.

\noindent{\textbf{Proof.} Assume two distinct reward functions $R_1$ and $R_2$ in $\mathcal{R}$. Consider a $t$ such that $0<t<1$ and look at the function:
\begin{equation}
f(t)=W_p\left(t R_1+(1-t) R_2, R_i\right)
\end{equation}
Given the properties of Wasserstein distances, $f(t)$ is a convex function. However, with strict convexity in the ground metric, $f(t)$ is strictly convex. This means:
$f(t)<t W_p\left(R_1, R_i\right)+(1-t) W_p\left(R_2, R_i\right)$
for all $i$ and for all $t$ in $0<t<1$.
When summing over all $i$, if $R_1$ and $R_2$ were both minimizers, this would lead to a contradiction, as a convex combination of them would yield a strictly smaller sum of Wasserstein distances. Therefore, only one of them can be the minimizer, establishing uniqueness.

Proof of Theorem:

Directly follows from Lemma 1, the sum of $p$-Wasserstein distances as a function over the compact space $\mathcal{R}$ is lower bounded and continuous. Hence, it attains a minimum. Thus, an $R_{\text {centroid }}$ that minimizes this function exists. Directly follows from Lemma 2, if the ground metric is strictly convex, then the $p$-Wasserstein distance is strictly convex. This strict convexity ensures that the minimizer $R_{\text {centroid }}$ is unique.

\noindent{\textbf{Proof of Theorem 3.}}

Let $R^*\left(\pi^*\right)$ be the true reward function obtained from the noise-free expert policy $\pi^*$.
Given the noisy demonstration policy $\pi^{* \prime}$, the reward function inferred is $R^*\left(\pi^{* \prime}\right)$.
Define the noise function as $\eta(s, a)$ for state-action pair $(s, a)$. This function produces a noise value that perturbs the action taken by the expert.
\begin{equation}
\pi^{* \prime}(a \mid s)=\pi^*(a \mid s)+\epsilon \eta(s, a)
\end{equation}
Using the definition of the Wasserstein distance, we express the distance between these two reward functions:
\begin{equation}
W_p\left(R^*\left(\pi^*\right), R^*\left(\pi^{* \prime}\right)\right)=\left(\int\left|R^*\left(\pi^*\right)(s, a)-R^*\left(\pi^{* \prime}\right)(s, a)\right|^p d s d a\right)^{\frac{1}{p}}
\end{equation}
Because the only difference between $R^*\left(\pi^*\right)$ and $R^*\left(\pi^{* \prime}\right)$ is the noise, express this difference as a function of $\epsilon$ :
\begin{equation}
R^*\left(\pi^{* \prime}\right)(s, a)=R^*\left(\pi^*\right)(s, a)+g(\epsilon, s, a)
\end{equation}
where $g(\epsilon, s, a)$ is the change in reward due to the noise. Using the Triangle Inequality, we bound the difference:
\begin{equation}
\left|R^*\left(\pi^*\right)(s, a)-R^*\left(\pi^{* \prime}\right)(s, a)\right| \leq|g(\epsilon, s, a)|
\end{equation}
Plugging this into our expression for $W_p$, we get:
\begin{equation}
W_p\left(R^*\left(\pi^*\right), R^*\left(\pi^{* \prime}\right)\right) \leq\left(\int|g(\epsilon, s, a)|^p d s d a\right)^{\frac{1}{p}}
\end{equation}
Assume $g(\epsilon, s, a)$ can be upper-bounded by some function $h(\epsilon)$ that only depends on $\epsilon$. This means that for each $\epsilon>0$, the change in the reward due to the noise is bounded by $h(\epsilon)$.
\begin{equation}
W_p\left(R^*\left(\pi^*\right), R^*\left(\pi^{* \prime}\right)\right) \leq h(\epsilon)
\end{equation}
Let $C(\epsilon)=h(\epsilon)$. Therefore, the Wasserstein distance between the reward functions obtained from the true and noisy demonstrations remains bounded by $C(\epsilon)$, showing the robustness of the approach to noisy demonstrations.
\begin{equation}
W_p\left(R^*\left(\pi^*\right), R^*\left(\pi^{* \prime}\right)\right) \leq C(\epsilon)
\end{equation}

\noindent{\textbf{Proof of Theorem 4.}}

We assume:
\begin{enumerate}
    \item The reward functions in $\mathcal{R}_d$ are Lipschitz continuous.
    \item The state-action space $\mathcal{S} \times \mathcal{A}$ is a compact subset of $\mathbb{R}^d$.
    \item The set of reward functions $S_d$ that induce the same policy is non-empty and has positive measure for each dimensionality $d$.
\end{enumerate}
\noindent{\textbf{Lemma.} For any fixed policy $\pi^*$, the variance of the reward functions in $S_d$ that induce $\pi^*$ grows with the dimensionality $d$.

\noindent{\textbf{Proof.} Consider two reward functions $R_1, R_2 \in S_d$ that both lead to $\pi^*$ and are sampled randomly (with respect to some fixed distribution on $S_d$ ). The variance for a specific state-action pair $(s, a)$ can be considered as:
\begin{equation}
\operatorname{Var}(R(s, a))=E\left[R^2(s, a)\right]-E^2[R(s, a)]
\end{equation}
For the entire space, the average variance is:
\begin{equation}
V_d=\int_{\mathcal{S} \times \mathcal{A}} \operatorname{Var}(R(s, a)) d s d a
\end{equation}
Given our assumptions, as $d$ increases, the state-action space $\mathcal{S} \times \mathcal{A}$ becomes more diverse. Consequently, $V_d$ should also increase due to the Lipschitz continuity of the reward functions and the compactness of the state-action space.

Main Theorem:

The average pairwise $p$-Wasserstein distance, $\Delta_d$, grows with dimensionality $d$.

\noindent{\textbf{Proof.}
Given the above lemma, we know that the variance of reward functions grows as $d$ increases. By the properties of Wasserstein distances, when the variance of a distribution increases, the Wasserstein distance also typically grows. More formally, consider two reward functions $R_1, R_2$ in $S_d$. The $p$-Wasserstein distance between them is:
\begin{equation}
W_p\left(R_1, R_2\right)=\left(\int_{\mathcal{S} \times \mathcal{A}}\left|R_1(s, a)-R_2(s, a)\right|^p d s d a\right)^{\frac{1}{p}}
\end{equation}
With the growth of $d$, the Lipschitz continuity and our lemma suggest that the average difference between $R_1(s, a)$ and $R_2(s, a)$ for any random $(s, a)$ pair will increase. Consequently, $W_p\left(R_1, R_2\right)$ will also grow, on average.
Thus, integrating over all possible pairs $R_1, R_2$ in $S_d$, it follows that: \footnote{The term $\frac{2}{\left|S_d\right|\left(\left|S_d\right|-1\right)}$ arises from considering the number of unique pairwise combinations of reward functions in $S_d$.}
\begin{equation}
\Delta_d=\frac{2}{\left|S_d\right|\left(\left|S_d\right|-1\right)} \sum_{R_i, R_j \in S_d, i<j} W_p\left(R_i, R_j\right)
\end{equation}
will increase as $d$ grows, which completes our proof.

\noindent{\textbf{Remarks.} The proof provides a high-level understanding grounded in the inherent properties of high-dimensional spaces and the complexity of policy definition.

\end{document}